\documentclass[10pt,journal,letterpaper]{IEEEtran}

\usepackage{amsmath,amssymb,amscd,bbm,amsthm,mathrsfs,dsfont}
\usepackage{algorithmic}
\usepackage{mdwmath}
\usepackage{mdwtab}
\usepackage{bm}
\usepackage{cite}
\usepackage{graphicx,psfrag}
\graphicspath{{Figures/}}
\IEEEoverridecommandlockouts

\usepackage{array}

\usepackage{booktabs}
\usepackage{indentfirst}
\usepackage{amsmath}

\begin{document}

\title{On Improving Energy Efficiency within Green Femtocell Networks: A Hierarchical Reinforcement Learning Approach}

\author{\IEEEauthorblockN{Xianfu Chen, Honggang Zhang, Tao Chen, Mika Lasanen, and Jacques Palicot}

\thanks{X. Chen, T. Chen, and M. Lasanen are with the VTT Technical Research Centre of Finland, P.O. Box 1100, FI-90571 Oulu, Finland (Email: \{xianfu.chen, tao.chen, mika.lasanen\}@vtt.fi).}

\thanks{H. Zhang is with the York-Zhejiang Lab for Cognitive Radio and Green Communications, and the Department of ISEE, Zhejiang University, Zheda Road 38, Hangzhou 310027, China (Email: honggangzhang@zju.edu.cn).}

\thanks{J. Palicot is with the SUPELEC/IETR, Avenue de la Boulaie, CS 47601, F-35576 Cedex-S\'{e}vign\'{e}, France (Email: jacques.palicot@supelec.fr).}
}

\maketitle

\begin{abstract}

One of the efficient solutions of improving coverage and increasing capacity in cellular networks is the deployment of femtocells. As the cellular networks are becoming more complex, energy consumption of whole network infrastructure is becoming important in terms of both operational costs and environmental impacts. This paper investigates energy efficiency of two-tier femtocell networks through combining game theory and stochastic learning. With the Stackelberg game formulation, a hierarchical reinforcement learning framework is applied for studying the joint expected utility maximization of macrocells and femtocells subject to the minimum signal-to-interference-plus-noise-ratio requirements. In the learning procedure, the macrocells act as leaders and the femtocells are followers. At each time step, the leaders commit to dynamic strategies based on the best responses of the followers, while the followers compete against each other with no further information but the leaders' transmission parameters. In this paper, we propose two reinforcement learning based intelligent algorithms to schedule each cell's stochastic power levels. Numerical experiments are presented to validate the investigations. The results show that the two learning algorithms substantially improve the energy efficiency of the femtocell networks.

\end{abstract}


\section{Introduction}

The insatiable desire for higher data rates and the requirement of ubiquitous internet access require a more dense deployment of base stations within the network cells. Whereas the traditional network infrastructures are less efficient, but it maybe not economical for the operators to make radical alternation to the current network architectures. Cellular networks are generally designed to provide large coverage and are not efficient in satisfying the need of ever increasing capacity-density. Therefore, cellular network deployment solutions based on femtocells are quite promising under this context \cite{Chandrasekhar08}. Due to the short transmit-receive distance property, femtocell techniques can greatly improve the indoor experience of the mobile users.

The escalation of energy consumption in wireless communications directly leads to the growth of greenhouse gas emission, which has been recognized as a major threat for environmental protection and sustainable development. Today, the increasingly rigid environmental standards have created an urgent need for green evolution in wireless communication networks\cite{Zhang12,Tao11}. In wireless cellular networks, the radio access section is the main source of energy consumption, accounting for up to more than $70\%$ of the total energy consumption.

To meet the challenges raised by the exponential growth in mobile services and energy consumption, it's crucial to increase the energy efficiency in wireless cellular networks. This paper addresses the energy efficiency problem in femtocell networks. The problem of energy-efficient spectrum sharing and power allocation in cognitive radio femtocells was studied in \cite{Xie12}, where a three-stage Stackelberg game model was formulated to improve the energy efficiency. In \cite{Ashraf}, Ashraf et al. proposed a novel energy saving procedure for the femtocell base station (FBS) to decide when to switch on/off. Hereinafter, we focus mainly on discussing the co-channel operation of femtocells with closed access. This is mainly due to the following reasons: 1) privacy concerns; 2) limited backhaul bandwidth; 3) no coordination between the macrocells and femtocells on spectrum allocation; 4) high requirements on mobile terminals.

On the other hand, in co-channel two-tier femtocell networks, the cross-tier/co-tier interference greatly restricts the overall network performance. Thus the interference cancelation in two-tier femtocell networks has become an active area of research. For the uplink transmission in two-tier femtocell networks, Chandrasekhar and Andrews \cite{Vikram} proposed a distributed utility-based signal-to-interference-plus-noise ratio (SINR) adaptation algorithm to alleviate the cross-tier interference at the macrocell from the co-channel femtocells. A Stackelberg game was formulated to study the resource allocation in two-tier femtocell networks, where the macrocell base station (MBS) protects itself by pricing the interference from femtocell users (FUs)\cite{RZhang}. In \cite{Jo09}, Jo et al. developed two interference mitigation strategies that adjust the maximum transmission power of FUs to control the cross-tier interference at a MBS. Regarding the downlink transmissions, Guruacharya et al. modeled the power allocation problem as a Stackelberg game to maximize the capacity of each station\cite{Guruacharya10}. And a macrocell beam subset selection strategy was used to reduce the cross-tier interference in two-tier femtocell networks in \cite{Park10}.

The unplanned deployment of femtocells results in unpredictable interference patterns. Therefore, the interference in this scenario can not be handled by means of centralized network scheduling, because the number and locations of femtocells are unknown. For such networking environment, the femtocells are most likely to be autonomous, which motivates using the idea of reinforcement learning (RL) \cite{Richard} for interference management. A realtime multi-agent RL algorithm that optimizes the network performance by managing the interference in femtocell networks was investigated in \cite{Giupponi}. Bennis et al. \cite{Bennis} developed a distributed learning scheme based on $Q$-learning to manage the femto-to-macrocell cross-tier interference in femtocell networks. Inspired by evolutionary game theory and machine learning, Nazir et al. \cite{Nazir10} proposed two intelligent mechanisms for interference mitigation to support the coexistence of macrocell and femtocells.

In this paper, we model the energy efficiency aspect of power allocation problem in femtocell networks as a Stackelberg learning game, i.e., leader-follower learning process, with the following characteristics: 1) the macrocells are considered to be the leaders, whereas the femtocells are considered to be the followers; 2) the leaders behave by knowing the response of the femtocells to their own strategy decisions; 3) given the leaders' decisions, the followers compete with each other. Learning is accomplished by directly interacting with the surrounding environment and properly adjusting the strategies according to the realizations of achieved performance. The solution of such a learning game is the Stackelberg equilibrium (SE). If no hierarchy\footnote{In this paper, the hierarchy means that the knowledge levels of the users are asymmetric.} exists during the learning procedure, the Stackelberg learning game reduces to the non-cooperative learning game, which is the scenario discussed in \cite{Qian}. Energy efficiency in wireless networks were studied using Stackelberg games in \cite{Lasaulce10, He11}.

Compared to the previous works, the main contributions of this paper are summarized as follows:
\begin{itemize}
  \item Firstly, for the energy efficiency problem in the femtocell networks, we propose a Stackelberg learning game for all users to jointly learn the optimal transmission strategies.
  \item Secondly, we develop a reinforcement learning based hierarchical power adaptation algorithm (\emph{RLHPA-I}) where the learning rule for FUs is based on each FU's private and incomplete information, and the MU behaves as the role of leader and learns the optimal transmission configuration by obtaining all FUs' strategy information; the trajectory of the learning dynamics is also investigated.
  \item Thirdly, in order to encourage the potential cooperation among the FUs, a second reinforcement learning based hierarchical power adaptation algorithm (\emph{RLHPA-II}) is further proposed, where the FUs' learn the optimal transmission strategies through conjectural beliefs over other competing FUs' stochastic behaviors; the convergence of the learning procedure is proved theoretically.
\end{itemize}

The rest of this paper is organized as follows. The next section presents the energy efficiency problem in femtocell networks and defines a Stackelberg game theoretic solution for the users' hierarchical behaviors. In Section III, a Stackelberg learning framework is proposed and the existence of SE is also investigated. Two reinforcement learning based algorithms are derived in Section IV and Section V. The numerical results are included in Section VI, verifying the validity and efficiency of the proposed algorithms. Finally, we present in Section VII a conclusion of this paper.

\section{Problem Formulation}

In this section, we first present the Stackelberg game formulation for the energy efficiency problem in femtocell networks. After that, the Stackelberg equilibrium of the proposed game is investigated.

\subsection{Stackelberg Game Formulation}

\begin{figure}
  \centering
  \includegraphics[width=0.33\textwidth]{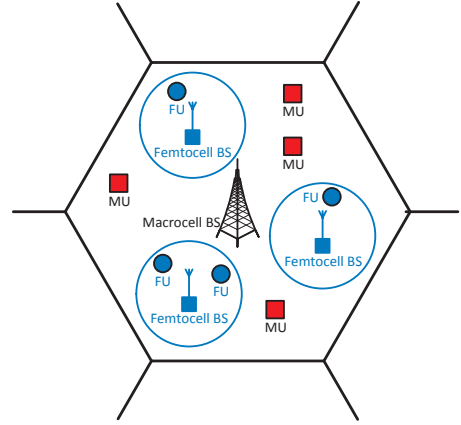}
  \caption{A typical femtocell network deployment (MBS: macrocell base station; MU: macrocell user; FBS: femtocell base station; FU: femtocell user).}
  \label{pict1}
\end{figure}

The femtocell network scenario we considered in this paper is illustrated in Fig. \ref{pict1}, where there exist multiple femtocells and macrocells. Each macrocell consisting of a MBS and multiple macrocell users (MUs), is underlaid with several co-channel FBSs. In each femtocell, there is one FBS providing service to femtocell users (FUs). Here we assume the closed independent policy \cite{Chandrasekhar08}, since private customers may prefer that kind of policy because of privacy concerns and limited backhaul bandwidth. Assuming identical distribution of femtocells in various neighboring macrocells, we focus our emphasis on the case of one representative macrocell for simplification, without loss of generality. Suppose $N$ femtocells $B_i(i\geq1)$ operate within the coverage of a macrocell $B_0$. Users of the same macrocell/femtocell adopt time division multiple access (TDMA) for data transmission, thus causing no interference within the same macrocell and femtocell. In the following, this paper mainly addresses the uplink transmissions for the distributed femtocells and the underlaid macrocell sharing a common spectrum band.

Let $i\in\mathcal{N}$ denote the scheduled user connected to its BS $B_i$, where $\mathcal{N}=\{0,1,\ldots,N\}$ refers to the index set of the MU and various FUs belonging to the same coexisting cellular region. Designate the transmission power level of user $i$ as $p_i\left(p_i^{\min}\leq p_i\leq p_i^{\max}\right)$, the SINR $\gamma_i$ of user $i$ received at $B_i$ is given by
\begin{equation}\label{sinr}
    \gamma_i^*\leq\gamma_i\left(p_i,\mathbf{p}_{-i}\right)=\dfrac{h_{i,i}p_i}{\sum_{j\in\mathcal{N}\setminus\{i\}}h_{j,i}p_j+\sigma^2},
\end{equation}
where $\gamma_i^*$ represents the minimum SINR requirement, $\sigma^2$ is the variance of background Additive White Gaussian Noise (AWGN), $\left\{h_{j,i}\right\}$ is the set of channel gains from user $j$ to $B_i$, and $\mathbf{p}_{-i}$ is a vector of power allocation for all users except user $i$, i.e., $\mathbf{p}_{-i}=\left(p_0,\ldots,p_{i-1}, p_{i+1},\ldots,p_N\right)$. In order to protect the MU's transmissions, we propose that the macrocell sets a power mask constraint for all femtocells \cite{Bohge09}, that is, the transmission power level of user $i\in\mathcal{N}\backslash\{0\}$ over the shared spectrum is constrained by
\begin{equation}\label{power_mask}
    p_i\leq p_{\footnotesize\mbox{mask}}.
\end{equation}
A power mask prescribes the maximum transmission power that the femtocells may use over the spectrum. From a practical viewpoint, this is much easier for the operators to manipulate for the scenarios where the number of active femtocells varies in time and space.

In the system under investigation, each user $i\in\mathcal{N}$ is selfish in the sense of its own energy efficiency, which can be expressed as\cite{Xie12, Xiong11}
\begin{equation}\label{EnergyEfficiency}
    \eta_i\left(p_i,\mathbf{p}_{-i}\right)=\dfrac{W\log_2\left(1+\gamma_i\left(p_i,\mathbf{p}_{-i}\right)\right)}{p_a+p_i},
\end{equation}
where $W$ is the spectrum bandwidth, and $p_a$ denotes the additional circuit power consumption of devices during transmissions (e.g., digital-to-analog converters, analog-to-digital converter, synthesizer,etc\cite{Cui04}) and is independent from the transmission power. Considering the QoS requirement in Eq. (\ref{sinr}), we define the utility function of user $i$ formally as
\begin{equation}\label{Utility}
    u_i\left(p_i,\mathbf{p}_{-i}\right)=\left\{
    \begin{array}{l@{\quad}l}
        \eta_i\left(p_i,\mathbf{p}_{-i}\right),&\mbox{if }\gamma_i\left(p_i,\mathbf{p}_{-i}\right)\geq\gamma_i^*;\\
        0, &\mbox{otherwise.}
    \end{array}
    \right.
\end{equation}
Eq. (\ref{Utility}) demonstrates interactions among the users. Each user $i$'s strategy is to choose the power level $p_i$ that maximizes its utility,
\begin{equation}\label{Op1}
    \max\limits_{p_i\in P_i}u_i\left(p_i,\mathbf{p}_{-i}\right),
\end{equation}
where $P_i=\left[p_i^{\min},\bar{p}_i^{\max}\right]$ is the strategy profile of user $i$, with $\bar{p}_i^{\max}=\min\left(p_i^{\max},p_{\footnotesize\mbox{mask}} \right)$. Particularly, $P_0=\left[p_0^{\min}, p_0^{\max}\right]$ for the MU.

In order to improve the energy efficiency, we introduce Stackelberg game \cite{Game} in the considered networking environment. Stackelberg game is a strategic game which consists of a leader and several followers competing with each other on some resources. Such a game formulation can be viewed as an intermediate scheme between the totally centralized power adaptation strategy and the non-cooperative strategy in \cite{Qian}. In this paper, the MU representing the whole MUs' coalition is modeled as the leader, while the FUs as the followers. Therefore, a distinct hierarchy exists among the users; and the leader plays the game by knowing the reaction function of the followers. The followers behave competitively, given the actions of the leader.

\subsection{Stackelberg Equilibrium Solution}

Game theory studies the rational interactions among the players. For the proposed Stackelberg game formulation, the SE describes an optimal strategy for the MU if all FUs always response by playing their Nash equilibrium (NE) strategies in the smaller sub-game. In order to investigate the existence of an SE, we first define $p_i^*$ to be the best response to $\mathbf{p}_{-i}$ if
\begin{align}
    u_i\left(p_i^*,\mathbf{p}_{-i}\right)\geq u_i\left(p_i,\mathbf{p}_{-i}\right),\forall p_i\in P_i.
\end{align}
User $i$'s best response to $\mathbf{p}_{-i}$ is denoted by $\mbox{\emph{BR}}_i(\mathbf{p}_{-i})$, maximizing its utility function subject to the power constraints. Let $\mbox{\emph{NE}}(p_0)$ be the NE strategy of the FUs if the MU chooses to play $p_0$, i.e.
\begin{align}
    \mbox{\emph{NE}}(p_0)=\mathbf{p}_{-0},\mbox{ if }p_i=\mbox{\emph{BR}}_i(\mathbf{p}_{-i}),\forall i\in\mathcal{N}\backslash\{0\}.
\end{align}

\textbf{Definition 1.} The strategy profile $\left(p_0^*,\mbox{\emph{NE}}(p_0^*)\right)$ is an SE if and only if
\begin{align}
    u_0\left(p_0^*,\mbox{\emph{NE}}(p_0^*)\right)\geq u_0\left(p_0,\mbox{\emph{NE}}(p_0)\right), \forall p_0\in P_0.
\end{align}

The following theorem establishes the existence of the SE.

\textbf{Theorem 1.} The SE always exists in our proposed Stackelberg game with the MU leading and the FUs following.

\emph{Proof}: In the proposed game formulation, it is not difficult to find that each FU $i\neq0$ strictly compete with other followers in a non-cooperative fashion, given the MU's action $\forall p_0\in P_0$. Therefore, a smaller non-cooperative power adaptation sub-game is formulated at the femtocell side $G=\big\langle p_0,\mathcal{N} \backslash\{0\},\{P_i\},\{u_i\}\big\rangle$. For a non-cooperative game, NE is a set of strategies, such that no player can benefit by changing its action unilaterally, assuming other players continue to use their current strategies. From the results in \cite{Xie12, Saraydar02}, there is at least one NE in the sub-game, since for $\forall i\in\mathcal{N}\backslash\{0\}$
\begin{enumerate}
  \item the strategy profile $P_i$ is a non-empty, convex, and compact subset of some Euclidean space $\mathfrak{R}^n$;
  \item $u_i$ is continuous in $(p_1,\ldots,p_{i-1},p_{i+1},\ldots,p_N)$ and quasi-concave in $p_i$.
\end{enumerate}
On the other hand, there is only one player at the macrocell side, and the best response strategy of the MU can be straightforwardly obtained through solving problem (\ref{Op1}). The above statement is thus proved. $\hfill\blacksquare$

We need to point out that in the Stackelberg game, the MU regards itself as the only leader and performs the Stackelberg strategy, and the FUs will act their best responses until reach the equilibrium $\left(p_0^*,\mbox{\emph{NE}}(p_0^*)\right)$. As the FUs, who are designated as the followers, are selfish, rational, and can not coordinate with each other. And they are going to play their best response strategies $\mbox{\emph{NE}}(p_0^*)$. By knowing this, the MU who is designated as the leader has to transmit with power level $p_0^*$ to maximize his utility function.

\section{Stackelberg Learning Framework}

In the Stackelberg learning game, each user in the network behaves as an intelligent agent, whose objective is to maximize its payoff. And the payoff is measured in utility function (e.g., Eq. (\ref{Utility})), which reflects the users' satisfaction of executing the strategy. The Stackelberg learning framework has two levels of hierarchy: 1) the MU learns to maximize its utility by knowing the response strategies of all FUs for each possible play; 2) given the strategy of the MU, the FUs play a non-cooperative learning game among each other. The game is played repeatedly to learn the optimal transmission strategies.

A strategy for user $i\in\mathcal{N}$ is defined to be a probability vector $\pi_i=\left(\pi_i(p_{i,1}),\ldots,\pi_i(p_{i,m_i})\right)\in\Pi_i$, where $\pi_i(p_{i,j_i})$ means the probability with which the user $i$ chooses action (transmission power) $p_{i,j_i}\in\mathcal{P}_i$, and $\Pi_i$ is the strategy set available to user $i$. Since each user can only choose a power level from a finite discrete set, $\mathcal{P}_i$ is assumed to be a finite set with dimension $m_i$. Then the expected utility function $U_i$ for user $i$ can then be expressed as follows
\begin{align}
    U_i\left(\pi_i,\bm\pi_{-i}\right)&=\mbox{E}\left[u_i|\mbox{user }j\mbox{ plays strategy }\pi_j,j\in\mathcal{N}\right]\nonumber\\
    &=\sum_{\textbf{p}\in\mathcal{P}}u_i\left(\textbf{p}\right)\prod_{s\in\mathcal{N}}\pi_s\left(p_{s,j_s}\right),
\end{align}
where $\bm\pi_{-i}=(\pi_0,\ldots,\pi_{i-1},\pi_{i+1},\ldots,\pi_N)$ is a vector of strategies for all other users, $\textbf{p}=\left(p_{0,j_0},\ldots,p_{N,j_N}\right)$ is the vector of actions chosen by all users, and $\mathcal{P}=\times_{i\in\mathcal{N}} \mathcal{P}_i$ is the space of all action vectors. An action suggests a power level performed by the user, and we use an action and a transmission power level interchangeably in the following discussions.

In the same way, we may have the following definition of SE in the proposed Stackelberg learning game.

\textbf{Definition 2.} For any stationary strategy\footnote{A strategy is said to be stationary, where $\pi_i=\left(\pi_i(1),\ldots,\pi_i(m_i)\right)$ is not changing with time during the stochastic learning process.} of the MU, $\pi_0\in\Pi_0$, the best-response strategies of all FUs define an NE strategy $\mbox{\emph{NE}}(\pi_0)$, i.e. $\mbox{\emph{NE}}(\pi_0)=\bm\pi_{-0}^*$, if
\begin{align}
    \pi_i^*=\arg\max_{\pi_i\in\Pi_i}U_i\left(\pi_i,\bm\pi_{-i}\right),\forall i\in\mathcal{N}\backslash\{0\}.
\end{align}
The MU's optimal strategy is then
\begin{align}
    \pi_0^*=\arg\max_{\pi_0\in\Pi_0}U_0\left(\pi_0,\mbox{\emph{NE}}(\pi_0)\right).
\end{align}
Together $(\pi_0^*,\mbox{\emph{NE}}(\pi_0^*))$ constitute a stationary strategy of SE for the Stackelberg learning formulation.

\textbf{Theorem 2.} For the proposed Stackelberg learning game, there exist a MU's stationary strategy and a FUs' NE strategy that form an SE.

Inspired by \cite{Yevgeniy}, we can prove Theorem 2 as follows.

\emph{Proof:} If the MU follows a stationary strategy $\pi_0\in\Pi_0$, then the Stackelberg learning game is simplified to be a $N$-player stochastic learning game for the FUs. It has been shown in \cite{Game} that every finite strategic-form game has a mixed strategy equilibrium. In other words, there always exists a stationary $\mbox{\emph{NE}} (\pi_0)$ that is best response for all the FUs in our formulation of the stochastic power adaptation process. The rest of the proof follows directly from the definition of SE, and is thus omitted for brevity. $\hfill\blacksquare$

Therefore, if we can construct an asymptotically (with time $t$) stationary strategy $\left\{\pi_i^t|i\in\mathcal{N}\right\}$ converging to the SE $(\pi_0^*,\mbox{\emph{NE}}(\pi_0^*))$, we will achieve the main goal of the Stackelberg learning power adaptation game in this paper. In the rest of this paper, we focus our emphasis on how to reach the optimal communication configuration through reinforcement learning approach.

\section{Reinforcement Learning based Hierarchical Power Adaptation-I (RLHPA-I)}

\subsection{Reinforcement Learning based Algorithm}

During the Stackelberg learning process, the MU behaves as the role of leader and knows the transmission strategy information of all FUs. Users with learning ability learn to maximize its individual expected utility function through repeated interactions with the surrounding networking environment. Among many different implementations of above adaptation mechanism, in this paper, we consider reinforcement learning, known as the so-called $Q$-learning \cite{Q-learning, Kianercy}, where the users' strategies are parameterized through $Q$-functions that characterize the relative expected utility of a particular power level. In $Q$-learning, users try to find the optimal $Q$-values in a recursive way. More specifically, let $Q_i^t\left(p_{i,j_i}\right)$ denote the $Q$-value of user $i$'s corresponding power level $p_{i,j_i}$ at time $t$. Then, after performing the transmission power level $p_{i,j_i}$ according to its strategy $\pi_i^t$ at time slot $t$, the $Q$-value is updated via the following rule
\begin{equation}\label{Q-Value-Update}
    Q_i^{t+1}\left(p_{i,j_i}\right)=\left(1-\alpha^t\right)Q_i^t\left(p_{i,j_i}\right)+\alpha^tU_i\left(p_{i,j_i},\bm\pi_{-i}^t\right),
\end{equation}
where $\alpha^t\in[0,1)$ is the learning rate, $\bm\pi_{-i}^t=(\pi_0^t,\ldots,$ $\pi_{i-1}^t,\pi_{i+1}^t,\ldots,\pi_N^t)$ is the vector of other users' strategies at time $t$, and
\begin{align}\label{Action-Expected-Utility}
    U_i\left(p_{i,j_i},\bm\pi_{-i}^t\right)=\sum_{\textbf{p}_{-i}\in\mathcal{P}_{-i}}u_i\left(p_{i,j_i},\textbf{p}_{-i}\right)\prod_{s\in\mathcal{N}\backslash \{i\}}\pi_s^t\left(p_{s,j_s}\right).
\end{align}
Here $\textbf{p}_{-i}=\left(p_{0,j_0},\ldots,p_{i-1,j_{i-1}},p_{i+1,j_{i+1}},\ldots,p_{N,j_N}\right)$ is a vector of actions chosen by all users except user $i$ over the action space $\mathcal{P}_{-i}=\times_{s\in\mathcal{N}\backslash\{i\}}\mathcal{P}_s$.

The tradeoff between \emph{exploration} and \emph{exploitation} is a challenge issue in stochastic learning process. The goal of \emph{Exploration} is to continually try new actions, while \emph{exploitation} aims to ``capitalize" on already established actions. One key feature of reinforcement learning is that it explicitly considers the \emph{exploration}$/$\emph{exploitation} in an integrated way, such that the users not only reinforce the actions they already know to be good but also explore new ones. In general, one deals with this problem through using a probabilistic method for choosing actions, e.g., $\epsilon$-greedy selection \cite{Eduardo} is an effective approach of balancing \emph{exploration} and \emph{exploitation}. One drawback, however, is that it might lead to globally suboptimal solution. Thus, we need to incorporate some way of exploring less-optimal actions.

An alternative solution is to vary the action probabilities as a graded function of the $Q$-values. The most common method is to use a Boltzmann distribution, that is, the probability of choosing transmission power level $p_{i,j_i}$ at time $t+1$ is given by
\begin{equation}\label{Boltzmann}
    \pi_i^t\left(p_{i,j_i}\right)=\frac{\exp\left(Q_i^t\left(p_{i,j_i}\right)/\tau_i\right)}{\sum_{p\in\mathcal{P}_i}\exp\left(Q_i^t(p)/\tau_i\right)},
\end{equation}
where $\tau_i$ is a positive parameter called the temperature and controls the exploration/exploitation tradeoff\cite{Richard}. A high temperature causes the action selection probabilities to be all nearly equal, while a low temperature results in big difference in selection probabilities for actions differ in their $Q$-values.

From Eq. (\ref{Q-Value-Update}) and Eq. (\ref{Action-Expected-Utility}), we can see that every user $i$'s updating rule depends on the strategies of other users. The MU who has the role of leader, can learn the optimal strategy according to Eq. (\ref{Q-Value-Update}) and Eq. (\ref{Boltzmann}). However, as the follower, each FU $i\in \mathcal{N} \backslash\{0\}$ can neither know other competing FUs' private strategy information $\bm\pi_{-(0,i)}^t=\left(\pi_1^t,\ldots,\pi_{i-1}^t, \pi_{i+1}^t,\ldots,\pi_N^t\right)$ nor the utility value $u_i(p_{i,j_i},\textbf{p}_{-i})$ before performing the action $p_{i,j_i}$. The only information it has is the MU's transmission parameters, i.e., the selected transmission power levels. Thus the updating rule for FU $i$ is transformed to
\begin{align}\label{Q-Value-Update-Follower}
    Q_i^{t+1}\left(p_{i,j_i}\right)=&\left(1-\alpha_f^t\right)Q_i^t\left(p_{i,j_i}\right)\nonumber\\
    &+\alpha_f^tU_i\left(p_{0,j_0},p_{i,j_i},\bm\pi_{-(0,i)}^t\right),
\end{align}
where $\alpha_f^t\in[0,1)$ is the learning rate for the FUs, and
\begin{align}\label{Action-Expected-Utility-Follower}
    &U_i\left(p_{0,j_0},p_{i,j_i},\bm\pi_{-(0,i)}^t\right)=\nonumber\\
    &\sum_{\textbf{p}_{-(0,i)}\in\mathcal{P}_{-(0,i)}}u_i\left(p_{0,j_0},p_{i,j_i},\textbf{p}_{-(0,i)}\right) \prod_{s\in\mathcal{N}\backslash\{0,i\}}\pi_s^t\left(p_{s,j_s}\right),
\end{align}
and $\textbf{p}_{-(0,i)}=\left(p_{1,j_1},\ldots,p_{i-1,j_{i-1}},p_{i+1,j_{i+1}},\ldots,p_{N,j_N}\right)$ is a vector of actions chosen by all FUs except FU $i$ over the action space $\mathcal{P}_{-(0,i)}=\times_{s\in\mathcal{N}\backslash\{0,i\}}\mathcal{P}_s$.

On the other hand, each FU $i\in\mathcal{N}\backslash\{0\}$ is able to compute the attainable utility $u_i(p_{i,j_i},\textbf{p}_{-i})$ with the feedback information (Eq. (\ref{sinr})) from its intended receiver. Under the Stackelberg learning framework, the MU behaves as the leader and makes decisions first. It's therefore assumed that the MU makes decisions every $T(>1)$ time slots, which is also defined as one episode. After each action is executed by the MU, all FUs repeatedly play the non-cooperative learning game during the episode. Suppose that the MU selects power level $p_{0,j_0}$ according to its strategy $\pi_0^k$ in episode $k$, the expected $U_i\big(p_{0,j_0},p_{i,j_i}, \bm\pi_{-(0,i)}^t\big)$ at time slot $t=(k-1)T+t_e$ $(t_e=1,\ldots,T)$ can be estimated using recursion in Eq. (\ref{Estimated-Action-Expected-Utility}),
\begin{figure*}
\begin{align}\label{Estimated-Action-Expected-Utility}
    \widetilde{U}_i^{t_e}\left(p_{0,j_0},p_{i,j_i}\right)=\left\{
    \begin{array}{l@{~}l}
        \dfrac{u_i\left(p_{0,j_0},p_{i,j_i},\textbf{p}_{-(0,i)}^t\right)-\widetilde{U}_i^{t_e-1}\left(p_{0,j_0},p_{i,j_i}\right)}
            {n_i^{k,t_e-1}\left(p_{i,j_i}\right)+1}+\widetilde{U}_i^{t_e-1}\left(p_{0,j_0}^k,p_{i,j_i}\right),&\mbox{if }p_{i,j_i}=p_{i,j_i}^t;\\
        \widetilde{U}_i^{t_e-1}\left(p_{0,j_0},p_{i,j_i}\right),&\mbox{otherwise.}
    \end{array}
    \right.
\end{align}
\hrule
\end{figure*}
where $\textbf{p}_{-(0,i)}^t=\big(p_{1,j_1}^t,\ldots,p_{i-1,j_{i-1}}^t,p_{i+1,j_{i+1}}^t,\ldots,p_{N,j_N}^t\big)$ is the vector of actions chosen by all other FUs except FU $i$ at time slot $t$, and $n_i^{k,t_e-1}\left(p_{i,j_i}\right)$ is the number of times when FU $i$ selects power level $p_{i,j_i}$ until time $t_e-1$ in episode $k$.

At any time slot $t_e\in\{1,\ldots,T\}$ in each episode $k$, each FU $i\in\mathcal{N}\backslash\{0\}$ is always supposed to know its own and the MU's actions. Substituting Eq. (\ref{Estimated-Action-Expected-Utility}) into Eq. (\ref{Q-Value-Update-Follower}), the $Q$-learning rule for FU $i$ can then be rewritten as
\begin{align}\label{Estimated-Q-Value-Update_Follower}
    Q_i^{t_e+1}\left(p_{i,j_i}\right)=\left(1-\alpha_f^{t_e}\right)Q_i^{t_e}\left(p_{i,j_i}\right)+\alpha_f^{t_e}\widetilde{U}_i^{t_e}\left(p_{0,j_0},p_{i,j_i}\right).
\end{align}
While the MU's learning algorithm resembles the standard single-agent $Q$-learning except for the fact that the expected utility is the utility accrued over one episode (i.e., $T$ time slots), that is,
\begin{align}\label{Estimated-Q-Value-Update_Leader}
    Q_0^{k+1}\left(p_{0,j_0}\right)=\left(1-\alpha_l^k\right)Q_0^k\left(p_{0,j_0}\right)+\alpha_l^k\mathbf{U}_0^k\left(p_{0,j_0}\right),
\end{align}
where $\alpha_l^k\in[0,1)$ is the learning rate for the MU, and
\begin{equation}\label{Averaged_Expected_Utility}
    \mathbf{U}_0^k\left(p_{0,j_0}\right)=\dfrac{1}{T}\sum\limits_{t_e\in\{1,\ldots,T\}}U_0\left(p_{0,j_0},\bm\pi_{-i}^{(k-1)T+t_e}\right).
\end{equation}
Accordingly, the strategy updates in Eq. (\ref{Boltzmann}) for the MU and the FUs are based on different time scales. Now we present the first reinforcement learning based hierarchical power adaptation algorithm for the Stackelberg learning game.\\
\rule{21pc}{0.1em}
\noindent\emph{RLHPA-I}

\noindent\rule{21pc}{0.06em}

\vspace{0.1cm}
\noindent\textbf{Initialization:}
\begin{enumerate}
\item[1)] $t=1(\mbox{such that } k=1)$, initialize $Q$-values $Q_i^t(p_{i,j_i})$ for each user $i\in\mathcal{N}$ and each action $p_{i,j_i}\in\mathcal{P}_i$.
\end{enumerate}

\noindent\textbf{Learning:}
\begin{enumerate}
        \item[2)] In episode $k$, the MU chooses action $p_{0,j_0}$ according to $\pi_0^k$ and broadcasts this information to all FUs in the network.
        \item[3)] Set $\widetilde{U}_i^{(k-1)T}(p_{0,j_0},p_{i,j_i})=0$ for each FU $i\in\mathcal{N}\backslash\{0\}$ and each action $p_{i,j_i}\in\mathcal{P}_i$. For $t=(k-1)T+1,\ldots,kT$, do.
          \item[(3.1)] FU $i$ selects an action $p_{i,j_i}$ according to $\pi_i^t$ and sends its relevant strategy information to the macrocell.
          \item[(3.2)] All users measure their SINR $\gamma_i$ with the feedback information of the intended receiver. If $\gamma_i\geq\gamma_i^*$, then $\eta_i\left(\textbf{p}\right)$ can be achieved; otherwise, the receiver can not receive correctly, thus obtains zero utility value.
          \item[(3.3)] The MU calculates $U_0\left(p_{0,j_0},\bm\pi_{-0}^t\right)$ according to Eq. (\ref{Action-Expected-Utility}).
          \item[(3.4)] All FUs update $\widetilde{U}_i^t\left(p_{0,j_0},p_{i,j_i}\right)$ basing on Eq. (\ref{Estimated-Action-Expected-Utility}).
          \item[(3.5)] All FUs update $Q$-values $Q_i^{t+1}\left(p_{i,j_i}\right)$ according to Eq. (\ref{Estimated-Q-Value-Update_Follower}).
          \item[(3.6)] All FUs update the strategies $\pi_i^{t+1}\left(p_{i,j_i}\right)$ according to Eq. (\ref{Boltzmann}).
          \item[(3.7)] Set $t=t+1$.
        \item[4)] The MU calculates $\mathbf{U}_0^k\left(p_{0,j_0}\right)$ according to Eq. (\ref{Averaged_Expected_Utility}).
        \item[5)] The MU updates $Q$-values $Q_0^{k+1}\left(p_{0,j_0}\right)$ according to Eq. (\ref{Estimated-Q-Value-Update_Leader}).
        \item[6)] The MU updates the strategies $\pi_0^{k+1}\left(p_{0,j_0}\right)$ according to Eq. (\ref{Boltzmann}).
        \item[7)] $k=k+1$.
\end{enumerate}

\noindent\textbf{End Learning}\\
\noindent\rule[0.1mm]{21pc}{0.1em}

The parameter $T$ decides the number of time slots that all FUs play the game before the MU updates its transmission strategy. Note that the updating rules of the MU and the FUs happen at different time scales. The FUs' $Q$-values are updated in every time slot whereas for the MU, the update happens only once in $T$ slots.

\subsection{Discussion of RLHPA-I}

From the definition of SE, it is clear that the convergence of \emph{RLHPA-I} to an SE requires that the MU's learning process converges to the optimal strategy while the FUs' stochastic behaviors converge to the corresponding NE under this optimal strategy. In this subsection, we discuss the conditions for such a convergence.

As already discussed, in the Stackelberg learning game we propose, the FUs behave as the followers in a smaller sub-game given the transmission strategy of the MU. In other words, for each strategy of the MU, the FUs have a multi-agent reinforcement learning problem in which the goal is to learn the NE of the game. In our algorithm \emph{RLHPA-I}, however, the FUs have independent learning processes that run simultaneously, with each one corresponding to each action of the MU. We use identical single-agent learning schemes for these processes and as already noted, the FUs maintain separate and private $Q$-values for each of these process. Each of these learning processes proceeds during $T$ time slots whenever the MU makes a decision according to its transmission strategy. That means each FU is also equipped with a standard single-agent reinforcement learning algorithm as the MU. Given that sufficient number of trails of the power levels have been executed, the FUs in our algorithm will converge to the NE responding to the MU's different transmission strategies.

The following Lemma by Szepesvari and Littman \cite{Szepesvari99} establishes the convergence of a general single-agent $Q$-learning process updated by a pseudo-contraction operator. Let $\bm{\mathbf{Q}}$ be the space of all $Q$-values.

\textbf{Lemma.} Assume that the learning rate $\alpha^t$ in Eq. (\ref{iteration}) satisfies the sufficient conditions of Theorem in \cite{Q-learning}, and the mapping $\mathcal{H}^t:\bm{\mathbf{Q}}\rightarrow\bm{\mathbf{Q}}$ meets the following condition: there exists a number $0<\beta<1$ and a sequence $x^t\geq 0$ converging to zero with probability (w.p.) $1$ as $t\rightarrow\infty$, such that $\left\|\mathcal{H}^tQ^t-\mathcal{H}^tQ^*\right\|\leq\beta\left\|Q^t-Q^*\right\|+x^t$ for all $Q^t\in\bm{\mathbf{Q}}$ and $Q^*=\mbox{E}\left[\mathcal{H}^tQ^* \right]$, then the iteration defined by
\begin{equation}\label{iteration}
    Q^{t+1}=\left(1-\alpha^t\right)Q^t+\alpha^t\left(\mathcal{H}^tQ^t\right),
\end{equation}
converges to $Q^*$ w.p. $1$.

\textbf{Theorem 3.} \emph{RLHPA-I} will always discover an SE strategy.

\emph{Proof:} We prove this by contradiction. Suppose that the process generated by Eq. (\ref{Boltzmann}) converges to a non-Stackelberg equilibrium. From previous discussion, we know that the long term behavior of \emph{RLHPA-I} converges to stationary points. This means that stationary points that are not SEs are stable, which contradicting Theorem 2. $\hfill\blacksquare$

Note that, unlike in the conventional single-agent reinforcement learning, in the considered Stackelberg learning problem, the MU's payoff value for performing a particular action is dependent on the outcome of a sub-game, which is played by the non-cooperative FUs in response to the MU's decision. When the FUs are in the process of learning their own transmission strategies, the outcomes of the smaller sub-games, and consequently, the utility values achieved by the MU, can typically be non-stationary. With non-stationary payoffs, the Lemma may not apply. In order to tackle this, we adopt an averaging procedure in our algorithm, as indicated by Eq. (\ref{Averaged_Expected_Utility}). At each updating step, the MU uses $\mathbf{U}_0^k\left(p_{0,j_0}\right)$, the averaged expected utilities from $T$ non-cooperative learning games of the FUs. This provides the MU with utilities that are good approximations of the payoffs corresponding to the outcomes of the non-cooperative sub-game.

\section{Reinforcement Learning based Hierarchical Power Adaptation-II (RLHPA-II)}

In order to promote potential cooperation among the competing FUs, we further propose a simple and intuitive rule that each FU links its own current transmission strategy to the other FUs' strategies. Such a rule reflects an awareness that there are strategic interactions during the learning procedure. FUs with such beliefs may not correctly perceive how the future strategies of their competitors depend on the past. In this section, we propose a conjecture model concerning the way in which the FUs react to each other.

\subsection{Conjecture Model}

Each FU $i\in\mathcal{N}\backslash\{0\}$ thinks any change in its current transmission strategy will induce other competing FUs to make well-defined changes in the corresponding time slot. Specifically, we need to estimate FU $i$'s expected contention measure at time slot $t=(k-1)T+t_e$, i.e., $b_{i}^t\big(\textbf{p}_{-(0,i)}\big)=\prod_{s\in\mathcal{N} \backslash\{0,i\}}\pi_s^t (p_{s,j_s})$ in Eq. (\ref{Action-Expected-Utility-Follower}), through a conjectural belief $\tilde{b}_i^t\big(\textbf{p}_{-(0,i)}\big)$, which is expressed as
\begin{equation}\label{Conjecture1}
    \tilde{b}_i^t\left(\textbf{p}_{-(0,i)}\right)=\overline{b}_i\left(\textbf{p}_{-(0,i)}\right)-\delta_i\left(\pi_i^t(p_{i,j_i})-\overline{\pi}_i(p_{i,j_i})\right),
\end{equation}
where the so-called reference points\cite{Tidball}, $\overline{b}_i\big(\textbf{p}_{-(0,i)}\big)$ and $\overline{\pi}_i(p_{i,j_i})$, are specific belief and probability, and $\delta_i>0$ is the belief factor. The reference points are considered as exogenously given. In other words, every FU $i$ believes that a change of $\pi_i^t(p_{i,j_i})- \overline{\pi}_i(p_{i,j_i})$ in its own strategy at time $t$ will induce a change of $\delta_i\left(\pi_i^t(p_{i,j_i})-\overline{\pi}_i(p_{i,j_i})\right)$ in the expected contention measure correspondingly related to the transmission strategies of other FUs. It's necessary to point out here that although FU $i$ may be aware that other FUs are subject to many influences on their strategies, when making its own decision, it is only concerned with other FUs' reactions to itself. That means FU $i$ does not take into account whether or not FU $s\left(s\in\mathcal{N}\backslash\{0,i\}\right)$ might react to changes in transmission strategy made by FU $v(v\in\mathcal{N}\backslash\{0,i,s\})$.

Among different possibilities of capturing the expected contention measure $b_i^t\big(\textbf{p}_{-(0,i)}\big)$, the linear model represented in Eq. (\ref{Conjecture1}) is the simplest form based on which one FU can model the impact of its changes in transmission strategy to the other competing FUs. In the non-cooperative learning process, as intelligent agents, the FUs learn when they modify the beliefs based on the new achievements. More specifically, we allow the FUs to revise their reference points according to their previous observations. That is, each FU $i$ sets $\overline{b}_i\big(\textbf{p}_{-(0,i)}\big)$ and $\overline{\pi}_i(p_{i,j_i})$ to be $b_i^{t-1}\big(\textbf{p}_ {-(0,i)}\big)$ and $\pi_i^{t-1}(p_{i,j_i})$. Eq. (\ref{Conjecture1}) then becomes
\begin{equation}\label{Conjecture}
    \tilde{b}_i^t\left(\textbf{p}_{-(0,i)}\right)=b_i^{t-1}\left(\textbf{p}_{-(0,i)}\right)-\delta_i\left(\pi_i^t(p_{i,j_i})-\pi_i^{t-1}(p_{i,j_i})\right).
\end{equation}
The conjecture model deployed by the FUs are based on the concept of reciprocity, which refers to the interaction mechanisms in which the FUs repeatedly interact when choosing the power level. If they realize that their probabilities of interacting with each other in the future is high, they will consider their influence on the strategies of other FUs, which is captured in the conjecture model by the positive parameter $\delta_i$. Otherwise, they will act myopically, which is the same learning process as in previous Section IV.

\subsection{Conjecture based Reinforcement Learning Scheme}

Following the previous analysis, the $Q$-learning rule for FU $i$ given by Eq. (\ref{Q-Value-Update-Follower}) is thus modified as Eq. (\ref{Estimated-Q-Value-Update_Follower2}).
\begin{figure*}
\begin{align}\label{Estimated-Q-Value-Update_Follower2}
    Q_i^{t_e+1}\left(p_{i,j_i}\right)=\left(1-\alpha_f^{t_e}\right)Q_i^{t_e}\left(p_{i,j_i}\right)+\alpha_f^{t_e}\sum_{\textbf{p}_{-(0,i)}\in\mathcal{P}_{-(0,i)}}u_i\left(p_{0,j_0},p_{i,j_i}, \textbf{p}_{-(0,i)}\right)\tilde{b}_i^{t_e}\left(\textbf{p}_{-(0,i)}\right)
\end{align}
\hrule
\end{figure*}
Therefore, we propose the second reinforcement learning based hierarchical power adaptation algorithm \emph{RLHPA-II} to discover the SE strategy. We may notice that the \emph{RLHPA-II} is quite similar to the \emph{RLHPA-I} except that the FUs update their $Q$-values based on Eq. (\ref{Estimated-Action-Expected-Utility}) in \emph{RLHPA-I}.

The detailed description of \emph{RLHPA-II} is given as follows.\\
\rule{21pc}{0.1em}
\noindent\emph{RLHPA-II}

\noindent\rule{21pc}{0.06em}

\vspace{0.1cm}
\noindent\textbf{Initialization:}
\begin{enumerate}
\item[1)] $t=1(\mbox{such that } k=1)$, initialize $Q$-values $Q_i^t(p_{i,j_i})$ for each user $i\in\mathcal{N}$ and each action $p_{i,j_i}\in\mathcal{P}_i$, and belief factors $\delta_i$ for each FU $i\in\mathcal{N}$.
\end{enumerate}

\noindent\textbf{Learning:}
\begin{enumerate}
        \item[2)] In episode $k$, the MU chooses action $p_{0,j_0}$ according to $\pi_0^k$, and the MBS broadcasts this information to all FUs in the network.
        \item[3)] For $t=(k-1)T+1,\ldots,kT$, do.
          \item[(3.1)] FU $i$ selects an action $p_{i,j_i}$ according to $\pi_i^t$ and sends its relevant strategy information to the macrocell.
          \item[(3.2)] All users measure their SINR $\gamma_i$ with the feedback information of the intended receiver. If $\gamma_i\geq\gamma_i^*$, then $\eta_i\left(\textbf{p}\right)$ can be achieved; otherwise, the receiver can not receive correctly, thus obtains zero utility value.
          \item[(3.3)] The MU calculates $U_0\left(p_{0,j_0},\bm\pi_{-0}^t\right)$ according to Eq. (\ref{Action-Expected-Utility}).
          \item[(3.4)] The MBS broadcasts strategy information $\bm\pi_{-0}^{t-1}$ to all FUs.
          \item[(3.5)] All FUs update $\tilde{b}_i^t\left(\textbf{p}_{-(0,i)}\right)$ basing on Eq. (\ref{Conjecture}).
          \item[(3.6)] All FUs update $Q$-values $Q_i^{t+1}\left(p_{i,j_i}\right)$ according to Eq. (\ref{Estimated-Q-Value-Update_Follower2}).
          \item[(3.7)] All FUs update the strategies $\pi_i^{t+1}\left(p_{i,j_i}\right)$ according to Eq. (\ref{Boltzmann}).
          \item[(3.8)] Set $t=t+1$.
        \item[4)] The MU calculates $\mathbf{U}_0^k\left(p_{0,j_0}\right)$ according to Eq. (\ref{Averaged_Expected_Utility}).
        \item[5)] The MU updates $Q$-values $Q_0^{k+1}\left(p_{0,j_0}\right)$ according to Eq. (\ref{Estimated-Q-Value-Update_Leader}).
        \item[6)] The MU updates the strategies $\pi_0^{k+1}\left(p_{0,j_0}\right)$ according to Eq. (\ref{Boltzmann}).
        \item[7)] $k=k+1$.
\end{enumerate}

\noindent\textbf{End Learning}\\
\noindent\rule[0.1mm]{21pc}{0.1em}

It's worth mentioning that during the learning process, every FU utilizes the other FUs' strategy information in previous time slot. Unlike \emph{RLHPA-I}, in algorithm \emph{RLHPA-II}, the FUs have multi-agent learning processes that relate to each other and run simultaneously.

\subsection{Theoretical Analysis of RLHPA-II}

Next, we concentrate on analyzing the convergence property of the \emph{RLHPA-II}. The algorithm results in a stochastic process of obtaining the vector of action selection probabilities, so we need to characterize the long-term behaviors of all users. Along with the discussion in Section IV-B, it only leaves us to prove the convergence of FUs' stochastic behavior in each episode $k$, given that $T$ is large enough. For an $N$-FU stochastic learning game, we define the operator $\mathcal{H}^{t_e}$ as follows.

\textbf{Definition 3.} Let $Q^{t_e}=\left(Q_1^{t_e},\ldots,Q_N^{t_e}\right)$, where $Q_i^{t_e}\in\bm{\mathbf{Q}}_i$ for $i\in\mathcal{N}\backslash\{0\}$, and $\bm{\mathbf{Q}}=\prod_{i\in \mathcal{N}\backslash\{0\}}\bm{\mathbf{Q}}_i$. $\mathcal{H}^{t_e}:\bm{\mathbf{Q}}\rightarrow\bm{\mathbf{Q}}$ is a mapping on the complete metric space $\bm{\mathbf{Q}}$ into $\bm{\mathbf{Q}}$, $\mathcal{H}^{t_e}Q^{t_e}=\left(\mathcal{H}^{t_e}Q_1^{t_e},\ldots,\mathcal{H}^{t_e}Q_N^{t_e}\right)$, where
\begin{align}
    &\mathcal{H}^{t_e}Q_i^{t_e}\left(p_{i,j_i}\right)=\nonumber\\
    &\sum_{\textbf{p}_{-(0,i)}\in\mathcal{P}_{-(0,i)}}u_i\left(p_{0,j_0},p_{i,j_i},\textbf{p}_{-(0,i)}\right) \tilde{b}_i^{t_e}\left(\textbf{p}_{-(0,i)}\right).
\end{align}

Then we proceed to prove that $Q^*=E\left[\mathcal{H}^{t_e}Q^*\right]$.

\textbf{Proposition 1.} For an $N$-FU stochastic game,
\begin{align}
    Q^*=\mbox{E}\left[\mathcal{H}^{t_e}Q^*\right],
\end{align}
where $Q^*=\left(Q_1^*,\ldots,Q_N^*\right)$.

\emph{Proof:} Since for $\forall i\in\mathcal{N}\backslash\{0\}$
\begin{align}
    &Q_i^*\left(p_{i,j_i}\right)\nonumber\\
    &=\mbox{E}\left[u_i\left(p_{0,j_0},p_{i,j_i},\bm\pi_{-(0,i)}^*\right)\right]\nonumber\\
    &=\sum_{\textbf{p}_{-(0,i)}\in\mathcal{P}_{-(0,i)}}u_i\left(p_{0,j_0},p_{i,j_i},\textbf{p}_{-(0,i)}\right) \prod_{s\in\mathcal{N}\setminus\{0,i\}}\pi_s^*(p_{s,j_s}).
\end{align}
From the discussions in previous Section V-A, we have $\tilde{b}_i^*\big(\textbf{p}_{-(0,i)}\big)=\prod_{s\in\mathcal{N}\setminus\{0,i\}}\pi_s^*(p_{s,j_s})$. Thus,
\begin{align}
    Q_i^*(p_{i,j_i})=\mbox{E}\left[\mathcal {H}^{t_e}Q_i^*(p_{i,j_i})\right],
\end{align}
for all $p_{i,j_i}\in\mathcal{P}_i$. $\hfill\blacksquare$

We further define the distance between any two $Q$-values.

\textbf{Definition 4.} For any $Q, Q'\in\bm{\mathbf{Q}}$, we define
\begin{equation}
    \big\|Q-Q'\big\|\triangleq\max\limits_{i\in\mathcal{N}\backslash\{0\}}\max\limits_{p_{i,j_i}\in\mathcal{P}_i}\big|Q_i(p_{i,j_i})-Q_i'(p_{i,j_i})\big|.
\end{equation}

\textbf{Proposition 2.} $\mathcal{H}^{t_e}$ is a contraction mapping operator.

\emph{Proof:} According to Definition 3, we have Eq. (\ref{equ0}).
\begin{figure*}
\begin{align}\label{equ0}
    \big\|\mathcal{H}^{t_e}Q-\mathcal{H}^{t_e}Q'\big\|
    &=\max\limits_{i\in\mathcal{N}\backslash\{0\}}\max\limits_{p_{i,j_i}\in\mathcal{P}_i}\big|\mathcal{H}^{t_e}Q_i(p_{i,j_i})-\mathcal{H}^{t_e}Q_i'(p_{i,j_i})\big|\nonumber\\
    &=\max\limits_{i\in\mathcal{N}\backslash\{0\}}\max\limits_{p_{i,j_i}\in\mathcal{P}_i}\left|\sum_{\textbf{p}_{-(0,i)}\in\mathcal{P}_{-(0,i)}}
                \left[\tilde{b}_i\left(\textbf{p}_{-(0,i)}\right)-\tilde{b}_i'\left(\textbf{p}_{-(0,i)}\right)\right] u_i\left(p_{0,j_0},p_{i,j_i},\textbf{p}_{-(0,i)}\right)\right|
\end{align}
\hrule
\end{figure*}
Next, we discuss the item $\sum_{\textbf{p}_{-(0,i)}\in\mathcal{P}_{-(0,i)}}\big[\tilde{b}_i\big(\textbf{p}_{-(0,i)}\big)-\tilde{b}_i'\big(\textbf{p}_{-(0,i)} \big)\big]u_i\big(p_{0,j_0},p_{i,j_i},\textbf{p}_{-(0,i)}\big)$ in Eq. (\ref{equ0}). Due to the fact that the reference points are exogenously given and of common knowledge, we may have Eq. (\ref{equ1}).
\begin{figure*}
\begin{align}\label{equ1}
    &\sum_{\textbf{p}_{-(0,i)}\in\mathcal{P}_{-(0,i)}}\left[\tilde{b}_i\left(\textbf{p}_{-(0,i)}\right)-\tilde{b}_i'\left(\textbf{p}_{-(0,i)} \right)\right]u_i\left(p_{0,j_0},p_{i,j_i},\textbf{p}_{-(0,i)}\right)\nonumber\\
    &=-\sum_{\textbf{p}_{-(0,i)}\in\mathcal{P}_{-(0,i)}}\delta_i\left[\pi_i(p_{i,j_i})-\pi_i'(p_{i,j_i})\right]u_i\left(p_{0,j_0},p_{i,j_i},\textbf{p}_{-(0,i)}\right)
\end{align}
\hrule
\end{figure*}

Now, we need to concentrate on the item $\pi_i(p_{i,j_i})$. By applying Eq. (\ref{Boltzmann}), we have
\begin{equation}
    \pi_i\left(p_{i,j_i}\right)=\dfrac{\exp\left(Q_i\left(p_{i,j_i}\right)/\tau_i\right)}{\sum_{p\in\mathcal{P}_i}\exp\left(Q_i(p)/\tau_i\right)}.
\end{equation}
When $\tau_i$ is sufficiently large, we have
\begin{equation}
    \exp\left(Q_i\left(p_{i,j_i}\right)/\tau_i\right)=1+\dfrac{Q_i\left(p_{i,j_i}\right)}{\tau_i}+\varphi\left(\dfrac{Q_i\left(p_{i,j_i}\right)}{\tau_i}\right),
\end{equation}
where $\varphi\big(Q_i(p_{i,j_i})/\tau_i\big)$ is a polynomial of the order $O\big((Q_i(p_{i,j_i})/\tau_i)^2\big)$. It's then straightforward to derive
\begin{align}
    \sum_{p\in\mathcal{P}_i}\exp\left(Q_i(p)/\tau_i\right)=m_i+\sum_{p\in\mathcal{P}_i}\left[\dfrac{Q_i(p)}{\tau_i}+\varphi\left(\frac{Q_i(p)}{\tau_i}\right)\right].
\end{align}

It can be easily verified that
\begin{align}\label{equ2}
    \pi_i\left(p_{i,j_i}\right)=\dfrac{1}{m_i}+\dfrac{1}{m_i}\cdot\dfrac{Q_i\left(p_{i,j_i}\right)}{\tau_i}+\varrho\left(\left\{\frac{Q_i(p)}{\tau_i}\right\}_p\right),
\end{align}
where $\varrho\big(\{Q_i(p)/\tau_i\}_p\big)$ is a polynomial of order smaller than $O\big(\{Q_i(p_{i,j_i})/\tau_i\}_p\big)$. Note that the coefficient of the polynomial is independent of the $Q$-value. Similarly, we may obtain
\begin{align}\label{equ3}
    \pi_i'\left(p_{i,j_i}\right)=\dfrac{1}{m_i}+\dfrac{1}{m_i}\cdot\dfrac{Q_i'\left(p_{i,j_i}\right)}{\tau_i}+\varrho\left(\left\{\frac{Q_i'(p)}{\tau_i}\right\}_p\right).
\end{align}

Substituting Eq. (\ref{equ2}) and Eq. (\ref{equ3}) to Eq. (\ref{equ1}) establishes Eq. (\ref{equ4}).
\begin{figure*}
\begin{align}\label{equ4}
    &\sum_{\textbf{p}_{-(0,i)}\in\mathcal{P}_{-(0,i)}}\left[\tilde{b}_i\left(\textbf{p}_{-(0,i)}\right)-\tilde{b}_i'\left(\textbf{p}_{-(0,i)}
        \right)\right]u_i\left(p_{0,j_0},p_{i,j_i},\textbf{p}_{-(0,i)}\right)\nonumber\\
    &=-\dfrac{\sum\limits_{\textbf{p}_{-(0,i)}\in\mathcal{P}_{-(0,i)}}\delta_i u_i\left(p_{0,j_0},p_{i,j_i},\textbf{p}_{-(0,i)}\right)}{\tau_i}\cdot\dfrac{1}{m_i}\left[
        Q_i\left(p_{i,j_i}\right)-Q_i'\left(p_{i,j_i}\right)\right]+\varrho\left(\left\{\frac{Q_i'(p)}{\tau_i}\right\}_p\right)- \varrho\left(\left\{\frac{Q_i(p)}{\tau_i}\right\}_p\right)
\end{align}
\hrule
\end{figure*}
This means we can always take a sufficiently large $\tau_i$ such that Eq. (\ref{equ5}) is satisfied,
\begin{figure*}
\begin{align}\label{equ5}
    \left|\sum_{\textbf{p}_{-(0,i)}\in\mathcal{P}_{-(0,i)}}\left[\tilde{b}_i\left(\textbf{p}_{-(0,i)}\right)-\tilde{b}_i'\left(\textbf{p}_{-(0,i)}\right)\right]
     u_i\left(p_{0,j_0},p_{i,j_i},\textbf{p}_{-(0,i)}\right)\right|\leq\dfrac{\lambda_i}{m_i}\big|Q_i\left(p_{i,j_i}\right)-Q_i'\left(p_{i,j_i}\right)\big|
\end{align}
\hrule
\end{figure*}
where $0<\lambda_i<m_i$. This implies
\begin{align}
    \big\|\mathcal{H}^{t_e}Q&-\mathcal{H}^{t_e}Q'\big\|\nonumber\\
    &\leq\max\limits_{i\in\mathcal{N}\backslash\{0\}}\max\limits_{p_{i,j_i}\in\mathcal{P}_i}\dfrac{\lambda_i}{m_i}
            \big|Q_i\left(p_{i,j_i}\right)-Q_i'\left(p_{i,j_i}\right)\big|\nonumber\\
    &\leq\omega\big\|Q-Q'\big\|,
\end{align}
where $\omega=\max_{i\in\mathcal{N}\backslash\{0\}}\frac{\lambda_i}{m_i}$. It's obvious that $\omega<1$.

Therefore, $\mathcal{H}^{t_e}$ is a contraction mapping operator. This concludes the proof. \hfill$\blacksquare$

We can now present the main result in this section that the learning process induced by the \emph{RLHPA-II} in each episode converges.

\textbf{Theorem 4.} For each FU $i\in\mathcal{N}\backslash\{0\}$, regardless of any initial value chosen for $Q_i^0(p_{i,j_i})$, if the temperature $\tau_i$ is sufficiently large, the FUs' stochastic behaviors converge.

\emph{Proof:} The proof can be completed by directly applying Lemma, which establishes the convergence given two conditions. First, $\mathcal{H}^{t_e}$ is a contraction mapping operator, by Proposition 2. Second, the fixed point condition, $Q^*=E[\mathcal{H}^{t_e}Q^*]$, is ensured by Proposition 1. Therefore, the learning process expressed by Eq. (\ref{Estimated-Q-Value-Update_Follower2}) converges. \hfill$\blacksquare$

\section{Numerical Results}

We provide insight into the performance comparison of the both learning algorithms through numerical simulations. We consider a representative macrocell scenario where there are two FUs coexisting with one MU over a spectrum with bandwidth of $1$MHz. The minimum SINR targets of MU and FUs are assumed to be $3$dB and $5$dB, respectively. The noise of the measurement is according to a zero-mean Gaussian noise with the power of $\sigma^2=-110$dBm, and the additional circuit power consumption is $10$dBm for all users. The femtocells are uniformly distributed within a circle area centered at the MBS with radius of $500$m. The coverage radius of femtocell is $20$m. The channel gains are generated by a log-normal shadowing pathloss model, $h_{i,j}=d_{i,j}^{-n}$, where $d_{i,j}$ is the distance between user $i$ and BS $j$, and $n$ is the pathloss exponent. In simulation, $n$ is assumed to be 4.

The action set of transmission power levels for all users is $\{20,25,30\}$dBm. Each episode contains $T=100$ time slots. For simplicity, we suppose that the belief factors $\delta_i$ are all equal to $2$ in \emph{RLHPA-II}, $\forall i\in\mathcal{N}\backslash\{0\}$, i.e., the FUs have the same conjecture ability. Further, we use the following learning rates for the MU and the FUs,
\begin{equation}
    \alpha_l^k=\dfrac{\alpha_l^1}{\theta^k},~~\alpha_f^{t_e}=\dfrac{\alpha_f^1}{\theta^{t_e}},
\end{equation}
where $\alpha_l^1,\alpha_f^1\in[0,1)$ are the initial learning rates, and $\theta>1$ is a scalar and is set to be $1.1$ in our simulations.

\begin{figure}[!t]
  \centering
  \includegraphics[width=0.45\textwidth]{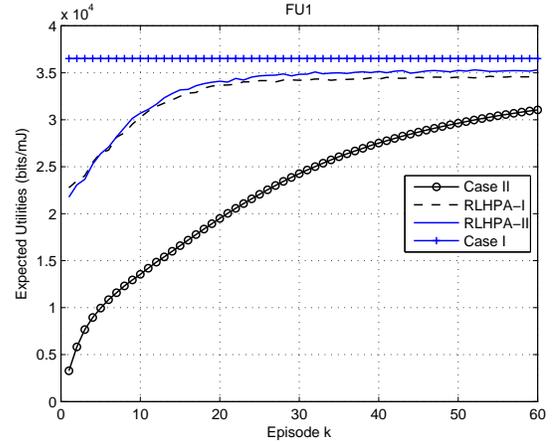}
  \caption{Learning process of the expected utilities for FU 1.}
  \label{pict2}
\end{figure}

\begin{figure}[!t]
  \centering
  \includegraphics[width=0.45\textwidth]{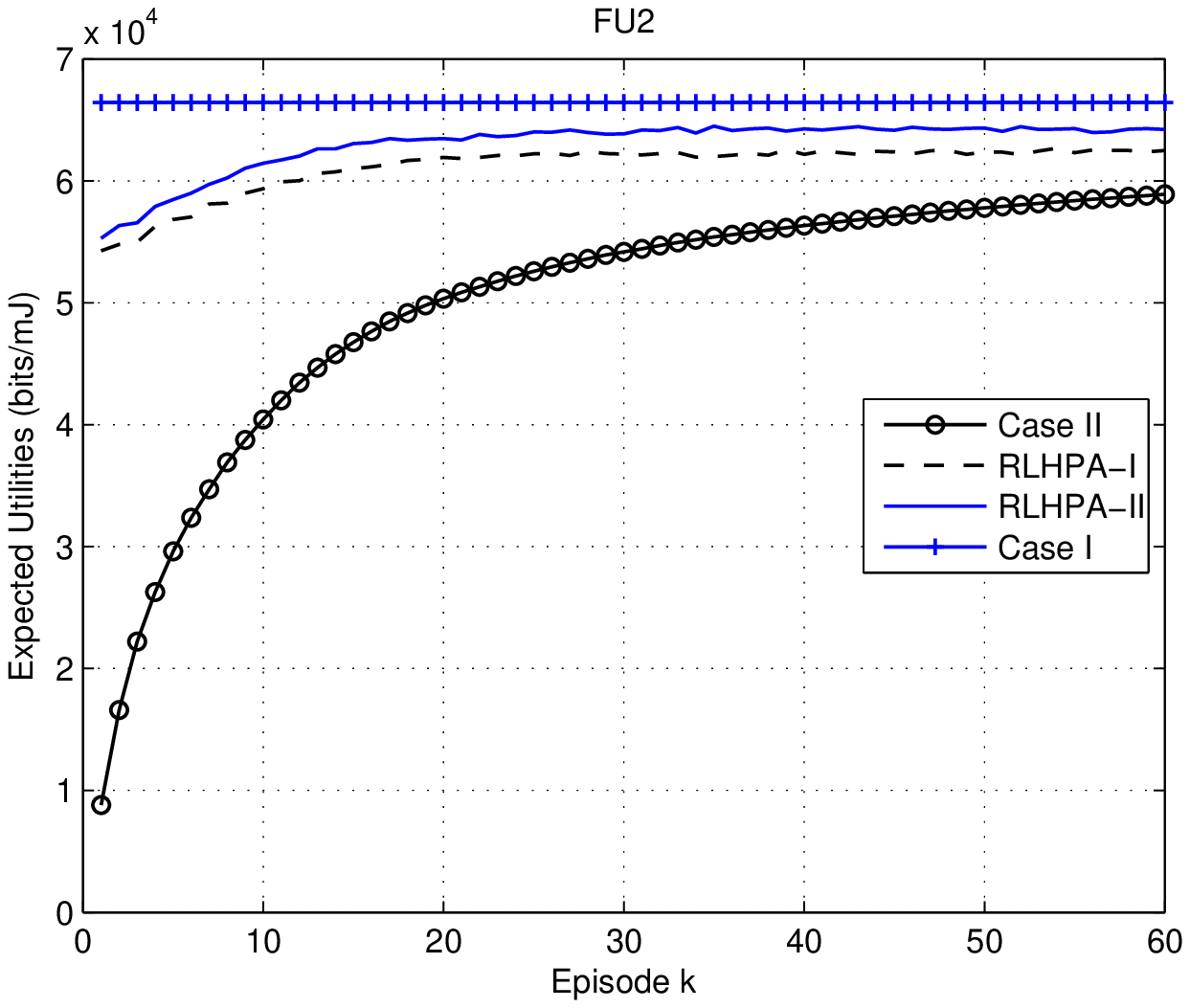}
  \caption{Learning process of the expected utilities for FU 2.}
  \label{pict3}
\end{figure}

\begin{figure}[!t]
  \centering
  \includegraphics[width=0.45\textwidth]{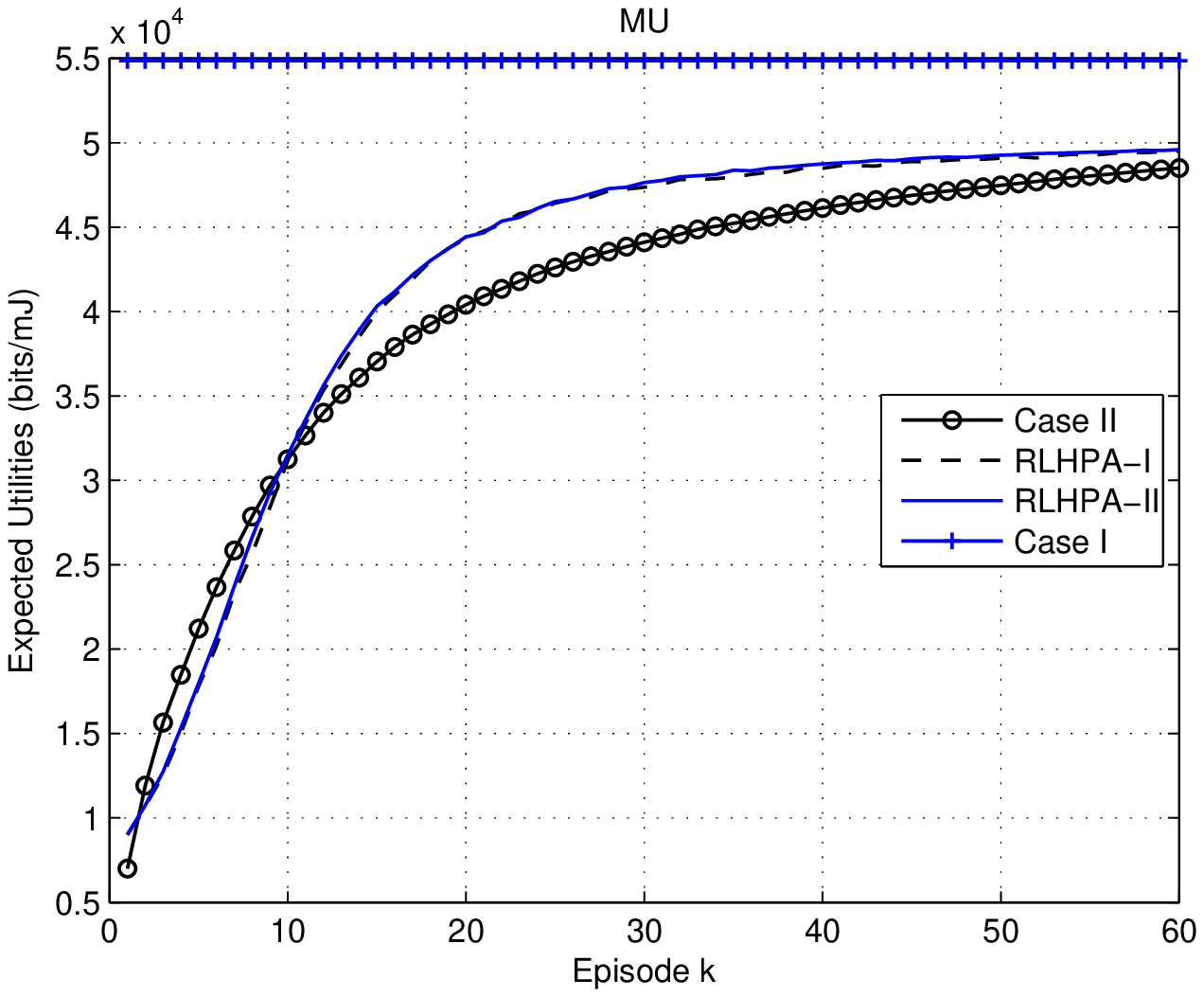}
  \caption{Learning process of the expected utilities for MU.}
  \label{pict4}
\end{figure}

The curves in Fig. \ref{pict2}, Fig. \ref{pict3} and Fig. \ref{pict4} show the learning process of expected utilities for each user in the network. The results are compared with
\begin{enumerate}
  \item The fully cooperative power allocation game with complete information exchange (Case I): each user knows all the utility functions and transmission power levels of other users in the network, and then the optimal utilities in the power allocation process can be obtained by each user according to Eq. (\ref{Op1}) through exhausted searching. This scenario is equivalent to the classic power control game in femtocell networks without the pricing schemes from macrocells as shown in \cite{Vikram}.
  \item The non-cooperative learning process of the power control game without any private information exchange (Case II): each user's transmission decisions in the learning process are self-incentive with myopic best response correspondence, which is the similar scenario discussed in \cite{Qian}.
\end{enumerate}
The first observation from our simulation results is that, whenever we generate random initial probability distributions of the power levels, the equilibrium state of the transmission strategies achieved by all users is independent of these initial values. That is, there exists an SE in the Stackelberg learning game, which confirms Theorem 2.

Secondly, we can find from the curves that the expected utilities of all users in the learning process will finally converge (or approach) to the equilibrium point in the complete cooperation case, and these simulation results validate the conclusions of Theorem 3 and Theorem 4. In addition, the proposed reinforcement learning based schemes both outperform the non-cooperative case, which is because for a Stackelberg learning game, knowing more can improve not only the leader's (MU) own utility, but also the utilities of the followers (FUs). Meanwhile, the \emph{RLHPA-II} can achieve better performance than \emph{RLHPA-I}, which is due to the fact that all FUs have the incentive to achieve better utilities thus behave reciprocally by exchanging transmission parameters in the previous time slot (indicated by Eq. (\ref{Conjecture})).

\begin{figure}[!t]
  \centering
  \includegraphics[width=0.45\textwidth]{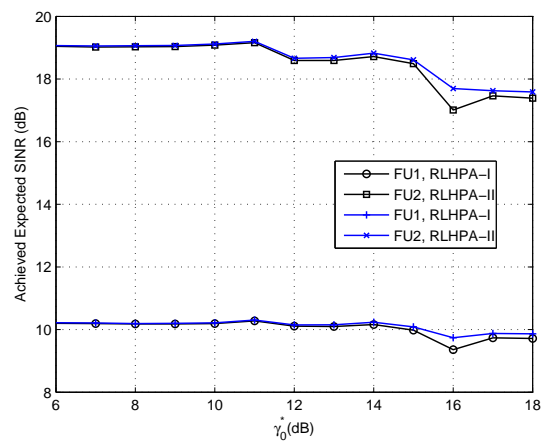}
  \caption{The expected SINRs for FUs versus $\gamma_0^*$.}
  \label{pict5}
\end{figure}

Fig. \ref{pict5} shows the expected SINRs of FUs using \emph{RLHPA-I} and \emph{RLHPA-II}, respectively, versus the macrocell's minimum QoS requirement $\gamma_0^*$. As expected, a higher $\gamma_0^*$ results in higher interference caused by the MU to the FUs, i.e., the achieved performances are degraded. Further, it can be observed that for the same $\gamma_0^*$, the expected SINRs of the FUs with \emph{RLHPA-II} is in general higher than that with \emph{RLHPA-I}. This is in accordance with our previous discussions. It is also worth mentioning that when $\gamma_0^*$ is sufficiently large, the expected SINRs of the FUs approach to zero for the two learning algorithms. This is because when $\gamma_0^*$ is sufficiently large, there is no femtocell active in the networks.

\section{Conclusion}

In this paper, energy efficiency is investigated for the uplink transmission in a spectrum-sharing-based two-tier femtocell network using stochastic learning theory combined with Stackelberg games. The Stackelberg learning framework is adopted to jointly study the utility maximization of the MU and FUs. Based on reinforcement learning, we propose two intelligent algorithms, namely, \emph{RLHPA-I} and \emph{RLHPA-II}, whose convergence properties have also been proven theoretically. Numerical experiments illustrate that the reciprocity-inspired \emph{RLHPA-II} converges more quickly and achieves better utility performance compared to \emph{RLHPA-I} and the non-cooperative learning scheme. This comes at the expense of obtaining more side strategy information at the FUs. Concludingly, both learning algorithms show the potential in improving the energy efficiency in the greener femtocell networks.


\begin{thebibliography}{29}

\bibitem{Chandrasekhar08}
V. Chandrasekhar, J. Andrews, and A. Gatherer, ``Femtocell networks: a survey," \emph{IEEE Commun. Mag.}, vol. 46, no. 9, pp. 59-67, Sep. 2008.

\bibitem{Zhang12}
J. Wu, S. Rangan, and H. Zhang, \emph{Green Communications - Theoretical Fundamentals, Algorithms and Applications}. CRC Press, Sep. 2012.

\bibitem{Tao11}
T. Chen, Y. Yang, H. Zhang, H. Kim, and K. Horneman, ``Network energy saving technologies for green wireless access networks," \emph{IEEE Wireless Commun.}, vol. 18, no. 5, pp. 30-38, Oct. 2011.


\bibitem{Xie12}
R. Xie, F. R. Yu, and H. Ji, ``Energy-efficient spectrum sharing and power allocation in cognitive radio femtocell networks," in \emph{Proc. INFOCOM}, Orlando, Florida USA, Mar. 2012.

\bibitem{Ashraf}
I. Ashraf, L. T. W. Ho, and H. Claussen, ``Improving energy efficiency of femtocell base stations via user activity detection," in \emph{Proc. WCNC}, Sydney, Australia, Apr. 2010.


\bibitem{Vikram}
V. Chandrasekhar and J. G. Andrews, ``Power control in two-tier femtocell networks," \emph{IEEE Trans. Wireless Commun.}, vol. 8, no. 8, pp. 4316-4328, Aug. 2009.

\bibitem{RZhang}
X. Kang, R. Zhang, and M. Motani, ``Price-based resource allocation for spectrum-sharing femtocell networks: A Stackelberg game approach," \emph{IEEE J. Sel. Areas Commun.}, vol. 30, no. 3, pp. 538-549, Apr. 2012.


\bibitem{Jo09}
H.-S. Jo, C. Mun, J. Moon, and J.-G. Yook, ''Interference mitigation using uplink power control for two-tier femtocell networks," \emph{IEEE Trans. Wireless Commun.}, vol. 8, no. 10, pp. 4906-4910, Oct. 2009.



\bibitem{Guruacharya10}
S. Guruacharya, D. Niyato, E. Hossain, and D. I. Kim, ''Hierarchical competition in femtocell-based cellular networks," in \emph{Proc. GLOBECOM}, Miami, FL, Dec. 2010.


\bibitem{Park10}
S. Park, W. Seo, Y. Kim, S. Lim, and D. Hong, ''Beam subset selection strategy for interference reduction in two-tier femtocell networks," \emph{IEEE Trans. Wireless Commun.}, vol. 9, no. 11, pp. 3440-3449, Nov. 2010.


\bibitem{Richard}
R. S. Sutton and A. G. Barto, \emph{Reinforcement Learning: An Introduction}. Cambridge, MA: MIT Press, 1998.


\bibitem{Giupponi}
L. Giupponi, A. M. Galindo-Serrano, and M. Dohle, ``From cognition to docition: The teaching radio paradigm for distributed \& autonomous deployments," \emph{Comput. Commun.}, vol. 33, no. 17, pp. 2015-2020, Nov. 2010.


\bibitem{Bennis}
M. Bennis, S. Guruacharya, and D. Niyato, ``Distributed learning strategies for interference mitigation in femtocell networks," in \emph{Proc. GLOBECOM}, Houston, Texas, USA, Dec. 2011.


\bibitem{Nazir10}
M. Nazir, M. Bennis, K. Ghaboosi, A. B. Mackenzie, and M. Latva-aho, ''Learning based mechanisms for interference mitigation in self-organized femtocell networks," in \emph{Proc. ASILOMAR}, Pacific Grove, CA, Nov. 2010.


\bibitem{Lasaulce10}
S. Lasaulce, Y. Hayel, R. El Azouzi, and M. Debbah, ''Introducing hierarchy in energy games," \emph{IEEE Trans. Wireless Commun.}, vol. 8, no. 7, pp. 3833-3843, Jul. 2009.


\bibitem{He11}
G. He, S. Lasaulce, and Y. Hayel, ''Stackelberg games for energy-efficient power control in wireless networks," in \emph{Proc. INFOCOM}, Shanghai, China, Apr. 2011.


\bibitem{Qian}
C. Long, Q. Zhang, B. Li, H. Yang, and X. Guan, ``Non-cooperative power control for wireless ad hoc networks with repeated games," \emph{IEEE J. Sel. Areas Commun.}, vol. 25, no. 6, pp. 1101-1112, Aug. 2007.


\bibitem{Bohge09}
M. Bohge, J. Gross, and A. Wolisz, ``Optimal power masking in soft frequency reuse based OFDMA networks," in \emph{Proc. EW}, Aalborg, Denmark, May 2009.


\bibitem{Xiong11}
C. Xiong, G. Y. Li, S. Zhang, Y. Chen, and S. Xu, ``Energy- and spectral-efficiency tradeoff in downlink OFDMA networks," \emph{IEEE Trans. Wireless Commun.}, vol. 10, no. 11, pp. 3874-3886, Nov. 2011.


\bibitem{Cui04}
S. Cui, A. J. Goldsmith, and A. Bahai, ``Energy-efficiency of MIMO and cooperative MIMO techniques in sensor networks," \emph{IEEE J. Sel. Areas Commun.}, vol. 22, no. 6, pp. 1089-1098, Aug. 2004.


\bibitem{Game}
D. Fudenberg and J. Tirole, \emph{Game Theory}. Cambridge, MA: MIT Press, 1992.


\bibitem{Saraydar02}
C. U. Saraydar, N. B. Mandayam, and D. J. Goodman, ``Efficient power control via pricing in wireless data networks," \emph{IEEE Trans. Commun.}, vol. 50, no. 2, pp. 291-303, Feb. 2002.



\bibitem{Yevgeniy}
Y. Vorobeychik and S. Singh, ``Computing Stackelberg equilibria in discounted stochastic games," in \emph{Proc. AAAI}, Toronto, Canada, Jul. 2012.

\bibitem{Q-learning}
C. J. C. H. Watkins and P. Dayan, ``$Q$-learning," \emph{Mach. Learn.}, vol. 8, no. 3-4, pp. 279-292, 1992.


\bibitem{Kianercy}
A. Kianercy and A. Galstyan, ``Dynamics of Boltzmann $Q$-learning in two-player two-action games," \emph{Phys. Rev. E}, vol. 85, no. 4, pp. 041145, April 2012.


\bibitem{Eduardo}
E. R. Gomes and R. Kowalczyk, ``Dynamic analysis of multiagent $Q$-learning with $\epsilon$-greedy exploration," in \emph{Proc. ICML}, Montreal, Canada, Jun. 2009.


\bibitem{Szepesvari99}
C. Szepesvari and M. L. Littman, ``A unified analysis of value-function-based reinforcement learning algorithm," \emph{Neural Comput.}, vol. 11 , no. 8, pp. 2017-2060, Nov. 1999.


\bibitem{Tidball}
A. Jean-Marie and M. Tidball, ``Adapting behaviors through a learning process," \emph{J. Econ. Behav. Organ.}, vol. 60, no. 3, pp. 399-422, Jul. 2006.

\end{thebibliography}
\end{document}